\pdfoutput=1

\documentclass[11pt]{article}
\usepackage[x11names]{xcolor}
\usepackage{acl}
\usepackage{amsmath}
\usepackage{times}
\usepackage{latexsym}
\usepackage{graphics}
\usepackage{graphicx}
\usepackage{color}
\usepackage{colortbl}
\usepackage{amssymb}
\usepackage{amsmath}
\usepackage{bbm}
\usepackage{booktabs}

\usepackage[T1]{fontenc}

\usepackage[utf8]{inputenc}

\usepackage{microtype}
\usepackage{multirow}
\usepackage{footmisc}
\usepackage{soul}
\usepackage{arydshln}

\usepackage{algpseudocode}
\usepackage{algorithm}
\usepackage{hyperref}
\usepackage{xspace}

\newcommand{\ie}{\textit{i.e.}}

\newcommand{\epm}{\textit{±}}

\title{InstructionNER: A Multi-Task Instruction-Based Generative Framework for Few-shot NER}

\author{Liwen Wang$^{1}$\thanks{\ \ The first two authors contribute equally. Weiran Xu is the corresponding author.} ,
Rumei Li$^{2*}$,
Yang Yan$^{1}$,
Yuanmeng Yan$^{1}$,
Sirui Wang$^{2}$,
Wei Wu$^{2}$,
Weiran Xu$^{1}$\\ 
$^1$Beijing University of Posts and Telecommunications, Beijing, China\\
$^{2}$Meituan Inc., Beijing, China\\
\texttt{\{w\_liwen,yanyang42,yanyuanmeng,xuweiran\}@bupt.edu.cn}\\
\texttt{\{lirumei,wangsirui,wuwei30\}@meituan.com}
}

\begin{document}
\maketitle
\begin{abstract}
Recently, prompt-based methods have achieved significant performance in few-shot learning scenarios by bridging the gap between language model pre-training and fine-tuning for downstream tasks. However, existing prompt templates are mostly designed for sentence-level tasks and are inappropriate for sequence labeling objectives. To address the above issue, we propose a multi-task instruction-based generative framework, named InstructionNER, for low-resource named entity recognition. Specifically, we reformulate the NER task as a generation problem, which enriches source sentences with task-specific instructions and answer options, then inferences the entities and types in natural language. We further propose two auxiliary tasks, including entity extraction and entity typing, which enable the model to capture more boundary information of entities and deepen the understanding of entity type semantics, respectively. Experimental results show that our method consistently outperforms other baselines on five datasets in few-shot settings.
\end{abstract}

\section{Introduction}
Pre-trained language models (PLMs) have achieved remarkable performance in natural language understanding and natural language generation tasks \cite{devlin-etal-2019-bert, lewis-etal-2020-bart, raffel2020exploring, NEURIPS2020_1457c0d6}. By constructing self-supervised pre-training tasks
, PLMs can learn plenty of syntactic, semantic, and even factual knowledge from a large amount of unlabeled corpus. As the increasing interests in exploring the knowledge learned by PLMs, prompt-based methods \cite{han2021pretrained} are proposed, which can reduce the gap between pre-training and 
fine-tuning process by reformulating the downstream tasks to pre-training-style tasks \cite{gao2021making,schick-schutze-2021-exploiting, sun2021nspbert}. Benefiting from the smaller task gaps and the inspiring prompt templates, the latent knowledge in PLMs can be effectively mined when performing downstream tasks. Thus, the prompt-based PLMs achieve significant improvements in the low-resource scenarios. 

\begin{figure}
    \centering
    \resizebox{.48\textwidth}{!}{
    \includegraphics{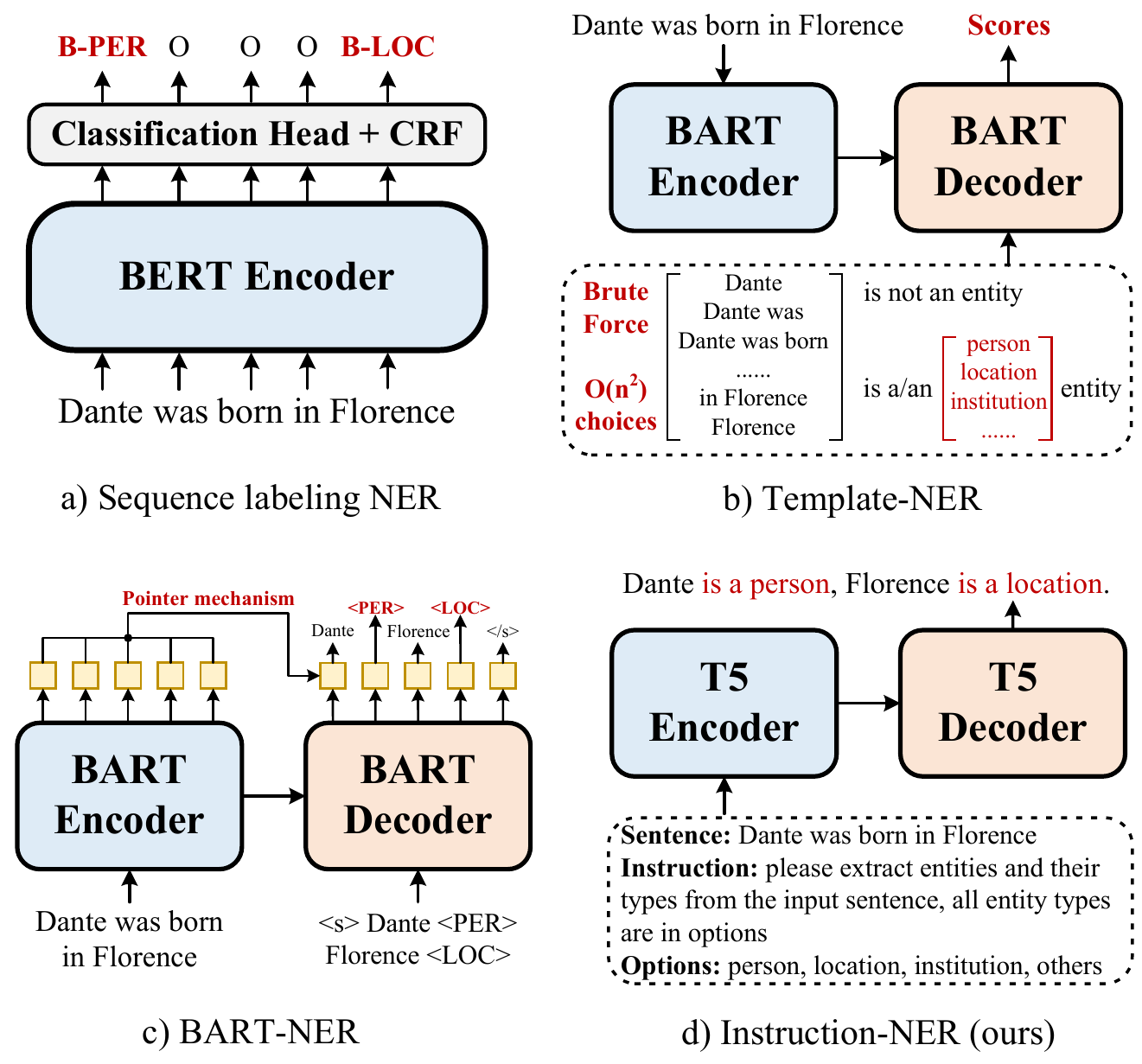}}
    \caption{Illustration of 4 different paradigms for solving NER task.}
    \label{fig:intro}
    \vspace{-0.5cm}
\end{figure}

The named entity recognition (NER) task, which requires the model to locate and classify named entities into predefined types, is generally formulated as a sequence labeling paradigm as shown in Figure \ref{fig:intro}a. NER serves as an important part for solving tasks such as information extraction \cite{ritter2012open}, question answering \cite{molla-etal-2006-named} and other language understanding problems \cite{guo2009named, gao-etal-2018-neural-approaches}. Unfortunately, the annotation resources for token labeling are often scarce and expensive in the real world. Thus, few-shot NER becomes a challenging but practical research problem and receives a lot of attention. Although prompt-based learning methods have achieved excellent performance in data-scarce scenarios, most of the existing templates-based methods are designed for sentence-level tasks and can be hardly fit for sequence labeling problems.
Therefore, the traditional NER paradigm needs to be reformulated to a more suitable way for PLMs, so that the prompt-learning methods can be applied to NER tasks. 

Recently, some works have started to reformulate NER tasks to sequence-to-sequence (seq2seq) tasks and integrate prompt-based methods. For instance, \citet{cui-etal-2021-template} proposes a template-based generative method (Template) which converts the NER task to span classification task and classifies span candidates in the form of cloze task at the decoding stage based on BART (shown in Figure \ref{fig:intro}b). While this method outperforms traditional sequence labeling baselines in few-shot scenarios as the introduction of prompt, it needs to enumerate all span candidates, which is inelegant and time-consuming. BARTNER \cite{yan-etal-2021-unified-generative} proposes a pointer-based seq2seq architecture, which converts NER sub-tasks to a unified sequence generation task and predicts entities from the input sentences and the corresponding type indexes (shown in Figure \ref{fig:intro}c). LightNER \cite{chen2021LightNER} introduces prompt-tuning to the attention mechanism of BARTNER and achieves promising improvement in low-resource scenarios. 
Inspired by TemplatedNER which designs specific prompt templates as decoder's input, and LightNER which introduces extra parameters as soft-prompts into the attention layer, we raise the question that, if we enrich the source sentence of generative PLMs with heuristic prompts, can this better stimulate the semantic knowledge learned in the pre-training stage and complete low-resource NER tasks?

To this end, we propose a multi-task instruction-based generative framework, named InstructionNER, for few-shot NER. 
Specifically, we reformulate the NER task as a natural language generation problem (shown in Figure \ref{fig:intro}d). For the source sentence, we design descriptive instructions to enable the model to understand different tasks \cite{wei2021finetuned}, and employ an option mechanism including all candidate entity categories as constraints of output space. Then, for inference, T5 \cite{raffel2020exploring} is required to generate the entity word and the corresponding type in the form of natural language, as we believe that unrestricted decoding would stimulate the latent knowledge of PLMs to complete entity extraction and recognition tasks to a larger extent.
Furthermore, we introduce two auxiliary tasks, named entity extraction (EE) task and entity typing (ET) task. EE requires the model to only decode the entity names and learn to better capture the boundary information. 
ET aims at only predicting the entity types and enhancing PLMs' understanding of type semantics.
Our contributions can be summarized as follows:
1) To fully leverage the knowledge in language models, we reformulate the NER task as a novel seq2seq problem, which integrates descriptive task instructions and answer options into the source sentence, then requires the model to predict entity names and types in natural language.
2) Moreover, we propose two auxiliary tasks which can enhance the ability to capture entity boundaries and the understanding of type semantics. 
3) Experiments on three NER benchmarks show the effectiveness of our proposed approach, especially in the data-scarce scenarios. In addition, we conduct a thorough analysis to show more characteristics of our approach.

\section{Related Works}

\subsection{Named Entity Recognition}


Currently, the most popular approach is to formulate NER as a sequence labeling task \cite{chiu-nichols-2016-named, strubell-etal-2017-fast, liu-etal-2019-gcdt, liu-etal-2021-noisy-labeled}, adding token-level classifiers or CRF \cite{ma-hovy-2016-end} on top of sentence encoders. This formulation is naturally suitable for flat NER, but can hardly fit for other subtasks \cite{ratinov-roth-2009-design, metke2016concept, strakova-etal-2019-neural, dai-etal-2020-effective}. 
Inspired by the recent success of pre-trained seq2seq models, BARTNER \cite{yan-etal-2021-unified-generative} reformulated all the three kinds of NER subtasks as a generation problem, and purposed a pointer-based framework to inference entities as well as their type index using BART \cite{lewis-etal-2020-bart}. Motivated by this new formulation, we treat NER as a natural language generation task, where the model is required to generate entity names and corresponding types in the form of natural language. Moreover, we employ T5 \cite{raffel2020exploring} instead of BART as our base model, as the pre-training task of T5 is to predict the sequence of corrupted tokens, which is more suitable for our formulation.

\subsection{Prompt-based Methods}

\paragraph{Prompt-based learning}

With the emergence of large pre-trained models like GPT3 \cite{NEURIPS2020_1457c0d6}, more and more researchers have begun to pay attention to a new fine-tuning paradigm, which is prompt-based learning. 
This new paradigm can make the best use of knowledge learned in the pre-training stage and thus achieve better performances in few-shot scenarios \cite{han2021pretrained}. 
Another line of work that tries to stimulate the potential of pre-trained models is instruction-based methods \cite{schick-schutze-2021-exploiting, mishra2021crosstask}. \citet{wei2021finetuned} proposed to fine-tune language models on a collection of task descriptions and answer options, for improving their abilities to respond to natural language instructions and better generalize to unseen tasks. Inspired by their work, we introduce designed instruction templates to describe each task and answer options, including candidate entity categories of the current dataset.

\paragraph{Prompt-based Few-shot NER}

To make full use of language models in low-resource NER, \citet{cui-etal-2021-template} proposed template-based BART, which treated original sentences as the source sequence, and statement templates filled by candidate spans as the target sequence. By introducing templates, this method outperforms traditional sequence labeling in few-shot scenarios, but it would be time-consuming to enumerate and classify all candidate spans. 
Following BARTNER, LightNER \cite{chen2021LightNER} incorporated continuous prompts into the self-attention matrix and constructed a semantic-aware answer space to replace label-specific layers. Different from their method, we inject task instructions and answer options into the source sentence, to stimulate more natural language knowledge from pre-trained models in few-shot settings.

\section{Methodology}

\begin{figure*}
    \centering
    \resizebox{\textwidth}{!}{
    \includegraphics{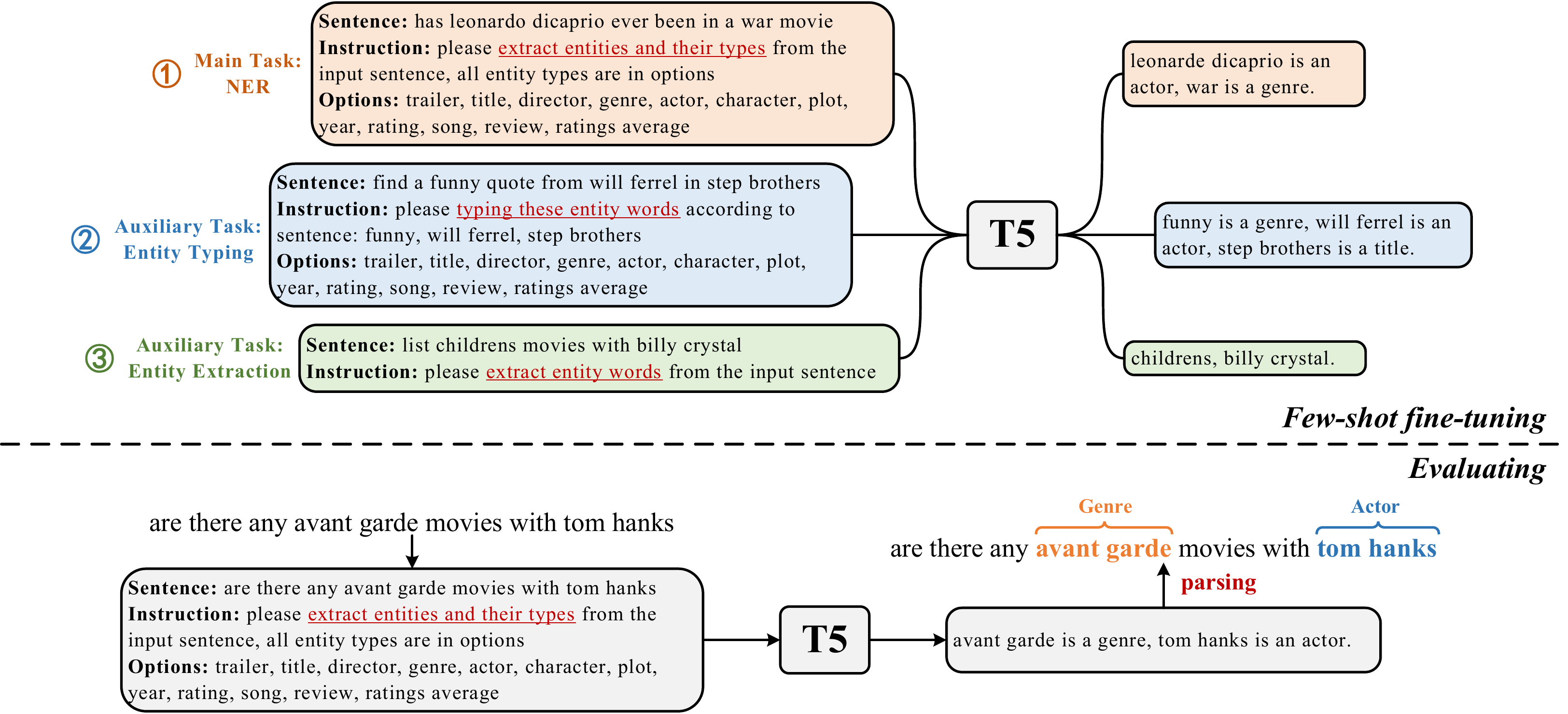}}
    \caption{The overall architecture of our proposed approach InstructionNER.}
    \label{fig:model}
    \vspace{-0.5cm}
\end{figure*}

In this section, we introduce the overall framework of our proposed InstructionNER. We first briefly describe the problem definition of NER, then we discuss how we convert NER to the seq2seq form so that it can be solved through T5. Next, we propose two auxiliary tasks, named entity extraction and entity typing, which help the model to better identify the entities for both the boundary and the type.
Finally, we introduce the parsing algorithm that converts the output of T5 (i.e., the word sequence) to the regular NER output (i.e., the triplet forms).

\paragraph{Problem Definition} The task of NER aims to recognize all entity occurrences $\boldsymbol{y}$ in a given sentence $\boldsymbol{x} = \{w_1, w_2, \dots, w_n\}$, where $n$ represents the length of $\boldsymbol{x}$. The $i$-th entity occurrence $y_i \in \boldsymbol{y}$ can be formulated as a triplet: $y_i = (l, r, t)$, where $l$ and $r$ indicate the left and right boundary indexes of the entity in the sentence $\boldsymbol{x}$, respectively. And $t \in \mathcal{T}$ indicates the entity type, where $\mathcal{T}$ is the full set of entity types. 

For simplicity, we use $\boldsymbol{x}_{l:r}$ to represent the span of $\boldsymbol{x}$ from the left boundary $l$ to the right boundary $r$ (inclusive), i.e., $\boldsymbol{x}_{l:r} = \{w_l, \dots, w_r\}$. Base on this, the entity occurrence $y_i = (l, r, t)$ indicates that the span $\boldsymbol{x}_{l:r}$ of sentence $\boldsymbol{x}$ is recognized as an entity of type $t$.

\paragraph{Solving NER through T5} \label{sec:method-t5} To better transfer and utilize the knowledge learned in pre-trained language models, we reformulate the NER task to the seq2seq form and solve it through fine-tuning T5 \cite{raffel2020exploring}, as shown in Figure \ref{fig:model}.

Specifically, for the main task (illustrated in the orange block in Figure \ref{fig:model}), each input consists of the following three fields:
\begin{itemize}
    \item \textbf{Sentence} - the source sentence $\boldsymbol{x}$.
    \item \textbf{Instruction} - the instruction tells the model which tasks the current sample belongs to. The model is trained to generate expected outputs that are consistent with the instruction. For the main NER task, the instruction is \textit{please extract entities and their types from the input sentence, all entity types are in options}.
    \item \textbf{Options} - all entity types $\mathcal{T}$, split by comma. This field acts as both a hint and a constraint to remind the model which entity types need to be recognized.
\end{itemize}

To stimulate the potential of the pre-trained model, we organize the output into a natural language form that naturally responses to the command of the input. Specifically, for the entity occurrence $(l, r, t)$, we convert it to the natural language form using the template: $\boldsymbol{x}_{l:r}$ is a/an $t$, and concatenate all converted entity occurrences together to make up an output sentence (using the comma as the separator and ending with the dot).

Specially, we have two strategies for filling the entity type $t$ in the template. One is naturally using the token or phrase of $t$ (e.g., using two tokens ``restaurant name'' to represent the entity type ``restaurant\_name''). Another is using synthetic tokens that represent $t$ (e.g., using a special token ``<restaurant\_name>'' to represent the entity type ``restaurant\_name''). The newly added special tokens would be appended to the token vocabulary of PLMs, and their embeddings are randomly initialized during the fine-tuning phase. We compare both strategies in our analysis experiments and get interesting results.

\paragraph{Auxiliary Tasks} To boost the performance in a more fine-grained level, we further design two auxiliary tasks, namely entity extraction and entity typing, which are exactly two fine-grained subtasks that compose the full NER task. We train the model jointly with these auxiliary tasks.

For the entity extraction task, the model is trained to extract the entity spans from the given sentence, but is not required to type them. We replace the instruction field with \textit{please extract entity words from the input sentence} to identify this task, and remove the options field, since there is no need for typing the extracted spans. Besides, the output should only contain the extracted spans, with ``is a/an $t$'' deleted.
The entity extraction task is a simplification of the main task. Guided by the instruction, the model only needs to extract the entities from sentences without focusing on the category information of the extracted entities, which helps a lot for improving the span F1 of entity extraction. The improvement of span F1 means that the ability to correctly extract entities from sentences is strengthened, and thus the accuracy on the NER main task is improved as well.

For the entity typing task, the model is trained to type the given entity occurrences in the sentence. Specifically, we replace the instruction field of the input sample with \textit{please typing these entity words according to the sentence: <the given entity occurrences>}, with other fields and the output same as the main task. 
In the training stage, as occurrences of entities in the sentence are given, we expect the model to focus more on capturing the category semantic information of entity occurrences. This encourages the model to generate more accurate category labels while maintaining the accuracy of entity occurrences generation in the main NER task, thus improving the performance of NER.

\vspace{-0.05cm}
\paragraph{Evaluation} The evaluation procedure of InstructionNER is shown in the half bottom of Figure \ref{fig:model}. Given the sentence $\boldsymbol{x}$, we first wrap it with the same template as we train the main NER task. Then we feed the input to the T5 model and obtain the generated output\footnote{Since T5 is a generative model, the beam search is used when evaluating.}. 

Once the output is obtained, we run the parsing algorithm to convert it to regular NER output as follows: 1) We split the output sentence by comma, and obtain multiple sub-sentences. 2) For each sub-sentence, we try to find the ``is a/an'' span. 3) If such a span is found, we intercept the prefix and suffix of the sub-sentence, and match them to the source sentence $\boldsymbol{x}$ to get predicted boundary indexes $\hat{l},\hat{r}$ and the set of entity types $\mathcal{T}$ to get predicted type $\hat{t}$, respectively. Then, we obtain a complete predicted triplet $(\hat{l},\hat{r},\hat{t})$. Once any step fails to match, we ignore this sub-sentence. After that, we compare predictions to the ground truth and use the F1 score as our main metrics.

\section{Experiments}
In this section, we conduct exhaustive few-shot experiments to evaluate the proposed InstructionNER framework and verify the effectiveness of the proposed multi-task strategy by comparing the performance of different implementations.

\subsection{Setup}

\paragraph{Datasets} Following \citet{chen2021LightNER,cui-etal-2021-template}, we use CoNLL-2003 as the rich-resource domain dataset, and following the settings in \cite{ziyadi2020examplebased,huang2020fewshot}, we use MIT Movie Review, MIT Restaurant Review \cite{liu-etal-2019-gcdt} and Airline Travel Information Systems (ATIS) \cite{HakkaniTr2016MultiDomainJS} dataset as low-resource datasets.

\paragraph{Baselines and Our Implementations}
In our experiments, we compare our methods with three types of NER methods and different variants of InstructionNER. 
\textbf{1) Sequence labeling Methods:} There are two methods, including LC-BERT and LC-BART, which apply the traditional sequence labeling method for named entity recognition with BERT and BART respectively.
\textbf{2) Template-based Methods:} A method proposed by \cite{cui-etal-2021-template}, which uses a BART-based seq2seq structure model to type all the enumerated spans by completing the human-designed template in line with the cloze task. We call this method Template for simplification.
\textbf{3) Generative seq2seq Methods:} BARTNER is a pointer-based seq2seq framework proposed by \cite{yan-etal-2021-unified-generative}, which convert the NER task to a unified sequence generation with a pointer mechanism. LightNER has a similar architecture with BART, but they introduce a prompt-guided attention mechanism, which is the only tuned module in the training process.
\textbf{4) Our Implementations} InstructionNER is trained without auxiliary task and InstructionNER{\scriptsize +ET/+EE/+ET, EE} is trained with the ET/EE/both auxiliary task(s), respectively.
    
\vspace{-0.3cm}

\paragraph{Settings}
In NER tasks, different from the classification tasks at the sentence level, each instance often has multiple entities. If we sample k sentences for each type of entity from the training set directly, the number of various entities in the sampled set will exceed k. In this paper, a greedy sampling strategy is utilized to guarantee the number of instances for each entity type is equal to k in the few-shot training set following \citet{yang2020simple}. If the number of occurrences of some entities is less than k in the training set, we will sample all these entities into the few-shot training set to keep in line with LightNER and Template. As the randomness of few-shot sampling, we repeat the sampling three times for each setting and report the average results and corresponding standard deviation.For fair comparisons with LightNER, Template, and BARTNER which employ BART-large as the base model, we also use the large version of T5 (T5-large) as the backbone model\footnote{If it is not specifically specified, the models used in our paper are all T5-Large (\url{https://huggingface.co/t5-large}).}. To keep the stability of the few-shot setting, we set the batch size to 2/4/8 for the 10/20/50 shot settings, respectively. In addition, we set the batch size to 16 for experiments with abundant samples. The learning rate of the Adam optimizer is set to 2e-5/5e-5, and the decoding beam search size is set to 2 to ensure more stable results and faster decoding speeds. More implement details are in appendix A.


\begin{table*}[htbp]
  \centering
  \resizebox{0.9\textwidth}{!}{
  \begin{tabular}{lcccccccccc}
    \toprule
    \multirow{2}{*}{Methods}&\multicolumn{3}{c}{MIT Movie}&\multicolumn{3}{c}{MIT Restaurant}&\multicolumn{3}{c}{ATIS} \\
    \cmidrule(lr){2-4}
    \cmidrule(lr){5-7}
    \cmidrule(lr){8-10}
    &10&20&50&10&20&50&10&20&50 \\
    \midrule
    \multicolumn{1}{l}{\textit{Baselines}} \\
    LC-BERT&25.2&42.2&49.6&21.8&39.4&52.7&44.1&76.7&90.7 \\
    LC-BART&10.2&27.5&44.2&6.3&8.5&51.3&42.0&72.7&87.5 \\
    Template&37.3&48.5&52.2&46.0&57.1&58.7&71.7&79.4&92.6 \\
    BARTNER$^*$&41.1&54.0&67.7&44.0&56.0&\underline{64.0}&\underline{77.7}&\underline{86.1}&\underline{93.4} \\
    LightNER&\underline{41.7}&\underline{57.8}&\underline{73.1}&\underline{48.5}&\underline{58.0}&62.0&76.3&85.3&92.8 \\
    \midrule
  
    \multicolumn{1}{l}{\textit{Our implementations}} \\

    InstructionNER&
    64.4 (\epm2.1)&70.0 (\epm0.3)&74.1 (\epm1.2)&58.7 (\epm1.2)& 65.5 (\epm1.4)& 71.2 (\epm1.1)&90.7(\epm 0.3)&93.0 (\epm 0.4)&95.1 (\epm 0.5) \\

    InstructionNER{\scriptsize +ET}&
    64.9 (\epm2.0)&
    \underline{\textbf{71.3} (\epm0.5)}&
    \underline{\textbf{75.6} (\epm0.8)}&
    
    59.1 (\epm0.3)&
    \underline{\textbf{67.2} (\epm0.8)}&
    \underline{\textbf{71.8} (\epm1.0)}&
    90.3 (\epm 0.1)&
    \underline{\textbf{93.1 (\epm 0.2)}}&
    95.3 (\epm 0.5)\\
    
    InstructionNER{\scriptsize +EE}&
    \underline{\textbf{66.1} (\epm2.5)}&
    70.8 (\epm0.2)&
    75.0 (\epm0.9)&
   
    \underline{\textbf{59.4} (\epm1.1)}&
    66.9 (\epm1.2)&
    71.5 (\epm0.7)&
    
    \underline{\textbf{90.8 (\epm 0.4)}}&
    \underline{\textbf{93.1 (\epm 0.2)}}&
    \underline{\textbf{95.4 (\epm 0.5)}}\\

    InstructionNER{\scriptsize +ET,EE}&
    65.6 (\epm3.0)&
    70.1 (\epm1.9)&
    74.7 (\epm0.3)&
   
    58.9 (\epm0.8)&
    66.1 (\epm0.9)&
    71.1 (\epm0.9)&
   
    90.6 (\epm 0.4) &
    93.0 (\epm 0.1)&
    95.3 (\epm 0.5)\\
    \bottomrule

   \end{tabular}}
  \caption{F1 score(\%) on three datasets under different shot settings. The value in brackets represents the standard deviation. \textbf{Bold} numbers indicate the best performance across baselines and our implementations, and \underline{underline} indicates the best performance in baselines or our implementations. `` * '' indicates our reproduction results.}
  \label{tab:main-results3}
  \vspace{-0.5cm}
  
\end{table*}

\paragraph{Metrics}
Similar to the baselines, we use F1 score as the evaluation metric for NER task. Besides, span F1 score is used to evaluate the entity extraction performance. The difference between F1 and span F1 is that, the true positive for F1 is the predictions with exactly matched entity extraction boundaries and correct entity type, while the true positive for span F1 only requires exactly matched entity extraction boundaries but ignores the entity type.

\subsection{Main Results}
Table \ref{tab:main-results3} shows the results of the proposed methods with various implementations and the competitive baselines. From the table, we have the following observations: 1) Under the setting of 10/20/50-shot, which are common in few-shot settings, our methods consistently outperform the compared baselines. Especially under the 10-shot setting, compared to LightNER, our best implementation achieves 24.4\%, 10.9\%, and 13.1\% F1 improvement on MIT Movie dataset, MIT Restaurant dataset, and ATIS dataset, respectively. 2) In our variants, the best implementation on MIT Movie dataset under the 10-shot setting (i.e., InstructionNER{\scriptsize +EE}) even outperforms the Template under 200-shot setting, 
which demonstrates the superiority of our model. 
3) When under the extreme data-scarce scenario (i.e., the 10-shot setting), we find the improvement over LightNER on MIT Movie (+24.4) is much more significant than that on MIT Restaurant (+10.9). We argue that the MIT Movie contains more entity types than MIT Restaurant, so that the generalization is much more difficult in MIT Movie under the 10-shot setting.  However, through converting the task into the natural language form with instruction and options, our approach shows powerful generalization ability, minimizing the impact of the insufficient data.

\subsection{Ablation Analysis}
Experimental results in Table \ref{tab:main-results3} show that our proposed InstructionNER has an outstanding ability for few-shot learning. To explore the influence of instruction-tuning and two auxiliary tasks on model performance, in this section, we do ablation analyses to compare and analyze our implementations in the bottom half of Table \ref{tab:main-results3}. We also conduct extra experiments to support our analysis.

\begin{figure}
    \centering
    \resizebox{.48\textwidth}{!}{
    \includegraphics{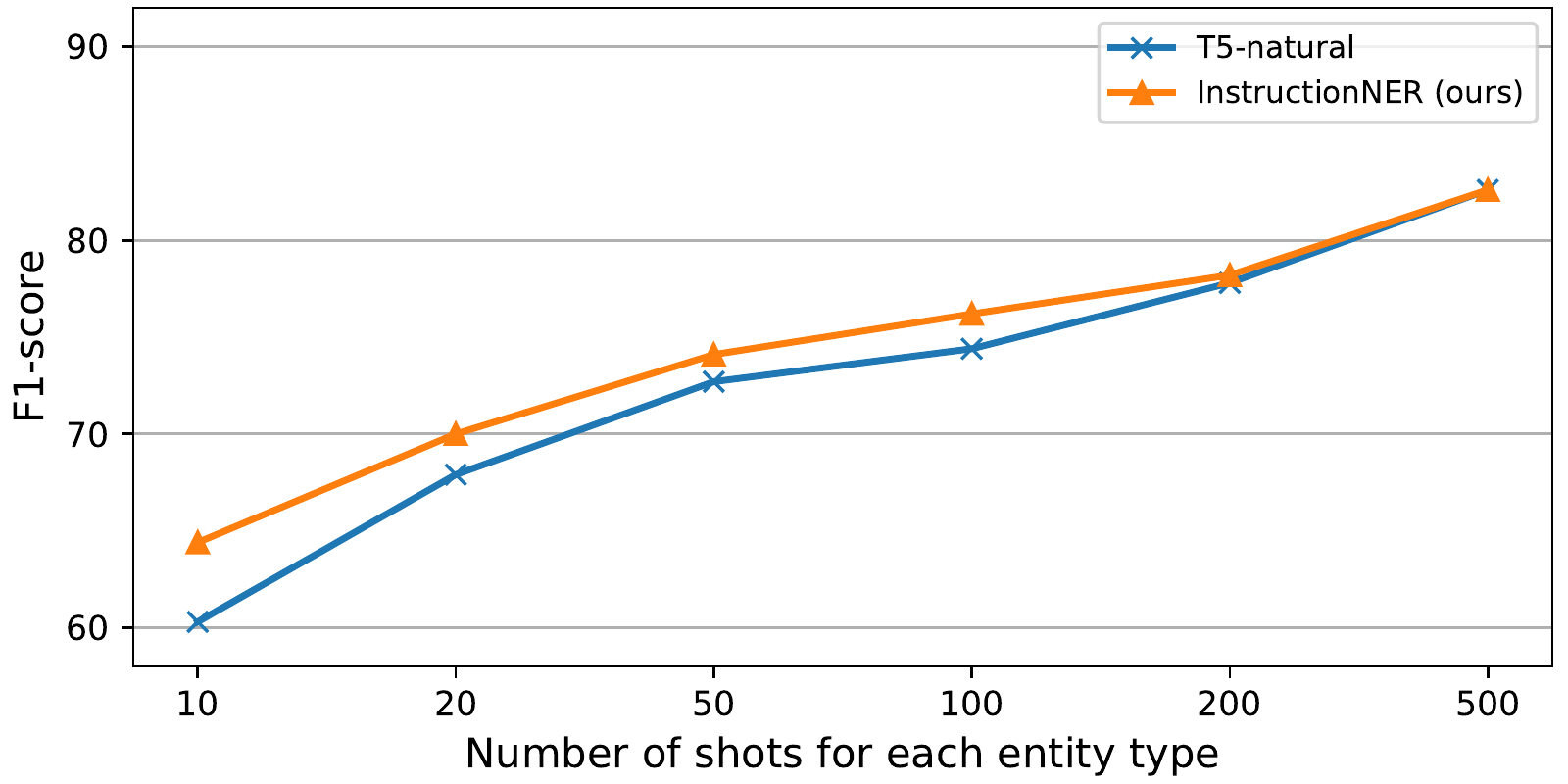}}
    \caption{Performance comparison between T5-natural and InstructionNER on.}
    \label{fig:instruction}
    \vspace{-0.5cm}
\end{figure}

\paragraph{Instruction} To demonstrate the effect of the instruction and the options, we conduct experiments that remove the instruction field and the options field when constructing the input (i.e., only input the raw sentence), but keep the output form unchanged (still the ``$\boldsymbol{x}_{l:r}$ is a/an $t$'' template for each entity occurrence $(l, r, t)$). Such a pattern is consistent with the pre-training phase of T5, where all tasks are converted to the seq2seq form. We call this baseline T5-natural.

In Figure \ref{fig:instruction}, we show the performance comparison between T5-natural and InstructionNER (both without auxiliary tasks). Generally, we can see that InstructionNER consistently outperforms the T5-natural among different data sizes, revealing the fact that instructions can effectively prompt the model and improve data efficiency under the few-shot setting.
On the other hand, we find the gap between the two models narrows as the data size increases. This also shows that the performance gain brought by the instruction and the options would decrease with more training data.

\begin{figure}
  \centering
  \resizebox{\columnwidth}{!}{
  \includegraphics{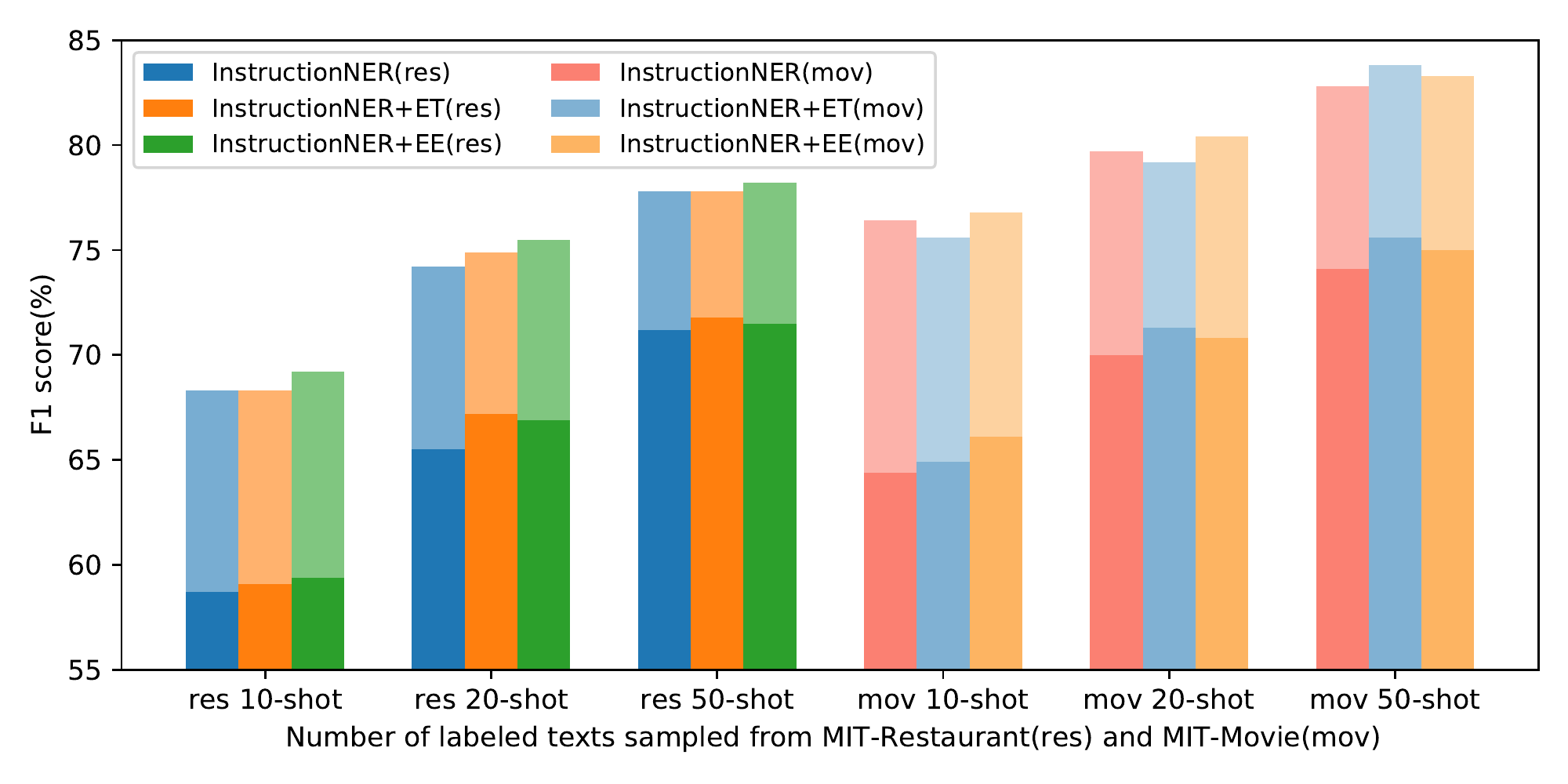}}
  \caption{F1 score(\%) on MIT Restaurant and MIT Movie datasets with different task combinations under 10/20/50 shot settings. The deep color indicates F1 score, while the full bar (light \& deep color) indicates span F1 score.}
  \label{fig:task-improve}
   \vspace{-0.5cm}
\end{figure}

\begin{table*}[t]
  \centering
  \resizebox{0.9\textwidth}{!}{
  \begin{tabular}{lcccccccccc}
    \toprule
    \multirow{2}{*}{Methods}&\multicolumn{3}{c}{CoNLL03}&\multicolumn{3}{c}{Ontonotes5.0}&\multicolumn{3}{c}{MIT Movie} \\
    \cmidrule(lr){2-4}
    \cmidrule(lr){5-7}
    \cmidrule(lr){8-10}
    &10&20&50&10&20&50&10&20&50 \\
    \midrule
    Template &
    44.8 (\epm 3.9)&
    64.3 (\epm 5.3)&
    72.8 (\epm 3.2)&
    24.1 (\epm 4.1)&
    36.5 (\epm 1.3)&
    43.9 (\epm 0.8)&
    38.7 (\epm 3.2)& 
    52.7 (\epm 1.4)& 
    61.1 (\epm 2.8)\\
    
    BARTNER &
    45.7 (\epm2.9)&
    62.4 (\epm5.4)&
    72.7 (\epm1.7)&
    20.9 (\epm 3.2)&
    33.2 (\epm 1.8)&
    44.6 (\epm 2.2)&
    41.1 (\epm 1.5)&
    54.0 (\epm 0.3)&
    67.7 (\epm 1.3)\\
    
    BART\_Instruction &
    15.4 (\epm 2.6)&
    31.6 (\epm 2.0)&
    39.0 (\epm 1.2)&
    12.4 (\epm 2.0)&
    21.7 (\epm 1.6)&
    34.4 (\epm 1.8)&
    21.3(\epm 4.4)&
    38.7(\epm 2.6)&
    51.3 (\epm 4.0)\\
    
     \midrule
  
    InstructionNER\_base &
    59.8 (\epm 2.2)&
    \underline{68.3} (\epm 0.2)&
    75.8 (\epm 2.3)&
    30.5 (\epm2.0)&
    39.6 (\epm2.3)&
    48.5 (\epm1.5)&
    58.9 (\epm4.2)&
    66.2 (\epm0.7)&
    72.8 (\epm1.8)\\
    
    InstructionNER\_v1.1&
    \underline{60.1} (\epm0.3)&
    \textbf{70.8} (\epm1.5)&
    \textbf{77.6} (\epm1.2)&
    \textbf{34.7} (\epm 2.0)&
    \textbf{42.0} (\epm 0.6)&
    \textbf{51.4} (\epm 0.9)&
    \textbf{66.4} (\epm2.3)&
    \underline{69.7} (\epm1.4)&
    \textbf{76.1} (\epm0.7)\\
    
    InstructionNER  &
    \textbf{61.3} (\epm2.0) &
    67.2 (\epm1.9)&
    \underline{76.3} (\epm1.0)&
    \underline{32.5} (\epm2.8)&
    \underline{41.4} (\epm 2.0)&
    \underline{50.2} (\epm 1.5)&
    \underline{64.4} (\epm2.1)&
    \textbf{70.0} (\epm0.3)& 
    \underline{74.1} (\epm1.2)    \\
    \bottomrule

  \end{tabular}}
  \caption{F1 score(\%) on CoNLL03, Ontonotes5.0, MIT Movie under 10/20/50-shot settings. The value in brackets represents the standard deviation.}
  \label{tab:new_res}
  \vspace{-0.5cm}
  
\end{table*}

\paragraph{Auxiliary Tasks}
As shown in Table \ref{tab:main-results3}, incorporating ET and EE auxiliary tasks can further boost the performance of InstructionNER under the low-resource NER settings. Specifically, under the typical low-resource settings of 10/20/50-shot, ET achieves 0.4/1.5/1.1 improvement on F1 score(\%) and EE achieves 1.2/1.2/0.6 improvement on F1 score(\%) on average in MIT Movie and MIT Restaurant dataset. To further explore the effectiveness of the two auxiliary tasks, we evaluate the span F1 score on MIT Restaurant and MIT Movie datasets. As shown in Figure \ref{fig:task-improve}, we have the following observations: 1) For InstructionNER{\scriptsize +ET}, the improvement on F1 score is more evident than the improvement on span F1 score. This phenomenon is consistent with the goal of the ET task that improves the performance of entity type classification when entity extraction is unchanged. 2) For InstructionNER{\scriptsize +EE}, the improvement of F1 score and span F1 score is approximately the same proportion, which indicates that incorporating EE task can extract more accurate entity boundaries and results in the improvement of NER task. Besides, we notice that the performance is dropped when we combine the ET and EE tasks.  We conjecture the reason is that the data size ratio between the main task and the auxiliary tasks is reduced to 1:2 when introducing ET and EE simultaneously, which may bring some noise to the main task\footnote{We discuss more details in Appendix D}.

\section{Analysis}

\subsection{Comparison of Different PLMs}
In this section, we conduct experiments on CoNLL03, Ontonotes5.0 and MIT-Movie, to make a detailed comparison between Template, BARTNER, BART\_Instruction (\ie, InstructionNER on BART),  InstructionNER\_base (\ie, InstructionNER on T5-base), InstructionNER\_v1.1 (\ie, InstructionNER on T5-v1.1\footnote{The released version 1.1 of T5-large, which is pre-trained on C4 only without mixing in the downstream tasks. See \url{https://huggingface.co/google/t5-v1_1-large} for more details.}), InstructionNER (the proposed methods). The experimental results are shown in \ref{tab:new_res}, and we would like to analyze the results from the following three perspectives:
\paragraph{Comparision Fairness}
Considering that T5-large is not only trained on unsupervised data (C4) but also on supervised tasks. There are some unfair comparisons. We replace T5 with T5 v1.1 and conduct experiments under InstructionNER framework. We find that the performance of the two implementations is close, which means that our improvements do not rely on the T5 version trained on supervised tasks.

\paragraph{BART vs. T5}
When observing the results of BART\_Instruction, we find that if BART is used as the backbone, the model performs very badly.
We think that the difference in performance is due to the pre-training task, BART needs to fully reconstruct the original input, while T5 only needs to predict the corrupted tokens, whose formulation is closer to our proposed Instruction-based NER task. Therefore, we employ T5 as the backbone in our experiments.

\paragraph{Model Scale}
When observing the results of InstructionNER\_base, we can see that although our proposed model uses T5 base as the backbone with fewer parameters, it also outperforms Template, which is based on BART-large. This proves that the Instruction-based framework is very effective for mining knowledge in text-to-text PLMs.

\subsection{Transfer Analysis}
In this section, we mainly focus on analyzing the performance of the proposed method in the low-resource cross-domain scenarios. Following \citet{chen2021LightNER} and \citet{cui-etal-2021-template}, we transfer the model trained on CoNLL03 (i.e., source domain) to MIT Restaurant (i.e., target domain) and randomly sample 10/20/50 instances per entity type as the continual training data. Three prototype-based methods, namely Neigh.Tag.\cite{wiseman-stratos-2019-label}, Example.\cite{ziyadi2020examplebased}, and MP-NSP\cite{huang2020fewshot}, respectively, are further introduced for comparison in the transfer scenarios.

As shown in Table \ref{tab:transfer-results}, the proposed InstructionNER outperforms the baselines in all three settings, which shows that our proposed model has good performance in the cross-domain few-shot NER task. Comparing the performance improvement of all methods after transferring, we find that the performance of LightNER has the most improvement. LightNER introduces the prompt-guided attention mechanism and this mechanism requires the model to update only a few parameters at every fine-tuning step, which prevents the model from overfitting in the source domain and strengthens the effect of cross-domain transfer. In addition, it is noticed that, in cross-domain transfer experiments, the performance of T5-natural is the worst and sometimes even has a negative impact. One possible reason is that T5-natural is an unrestricted seq2seq NER architecture, when the input text in the source domain differentiates a lot from the text in the target domain, the knowledge learned in the source domain is limited and even treated as noise, which has a negative impact. However, InstructionNER prompts the input and output of the model via instruction and options. Thus, when the model is transferred from the source domain to the target domain, the variance of options can help the model adapt to the new domain and improve the transfer performance of the generative NER framework.

\begin{table}[t]
  \centering
  \resizebox{0.9\columnwidth}{!}{
  \begin{tabular}{lccc}
    \toprule
    \multirow{2}{*}{Methods}&\multicolumn{3}{c}{MIT Restaurant} \\
    \cmidrule(lr){2-4}
    &10&20&50 \\
    \midrule
    \multicolumn{4}{c}{\textit{Baselines}} \\
    Neigh.Tag.&4.1&3.6&4.0 \\
    Example.&25.2&26.1&26.8 \\
    MP-NSP&46.1&48.2&49.6 \\
    LC-BERT&27.2(+5.4)&40.9(+1.5)&56.3(+3.6) \\
    LC-BART&8.8(+2.5)&11.1(+2.6)&42.7(-8.6) \\
    Template&53.1(+7.1)&60.3(+3.2)&64.1(+5.4) \\
    BARTNER$^*$&55.0(+11.0)&64.0(+8.0)&71.5(+7.5) \\
    LightNER&58.1(+9.6)&67.4(+9.4)&69.5(+7.5) \\
    \midrule
    \multicolumn{4}{c}{\textit{Our implementations}} \\
    T5-natural&55.8(+2.6)&63.6(+1.2)&68.6(-0.5) \\
    InstructionNER&\textbf{63.7(+5.0)}&\textbf{68.2(+2.7)}&\textbf{71.9(+0.7)} \\
    \bottomrule
    
  \end{tabular}}
  \caption{F1 score(\%) on MIT Restaurant dataset under transfer setting. The value in brackets represents the improvement compared to the in-domain results (i.e., results in Table \ref{tab:main-results3}). \textbf{Bold} numbers indicate the best performance. `` * '' indicates our reproduction results.}
  \label{tab:transfer-results}
  \vspace{-0.6cm}
  
\end{table}

\subsection{Label Semantic Analysis}
\label{sec:label_semantic_analysis}

As mentioned in Section \ref{sec:method-t5}, there are two strategies to fill the entity types in the template. One is using the natural language form, while the other regards type words as synthesized tokens and adding them to the vocabulary. Intuitively, InstructionNER can leverage the semantic information in label words to enhance the model in low-resource scenarios. In previous experiments, we use the setting of type words in the natural language form except for the experiments on ATIS\footnote{ATIS include 79 entity types and using natural language labels directly would make the input Instruction too long.}. In this section, we explore the influence of the two strategies on the performance of InstructionNER. We design multiple experimental scenarios from 10-shot to full-supervised on MIT Movie dataset and MIT Restaurant dataset to explore the two strategies.

\begin{figure}
  \centering
  \resizebox{0.95\columnwidth}{!}{
  \includegraphics{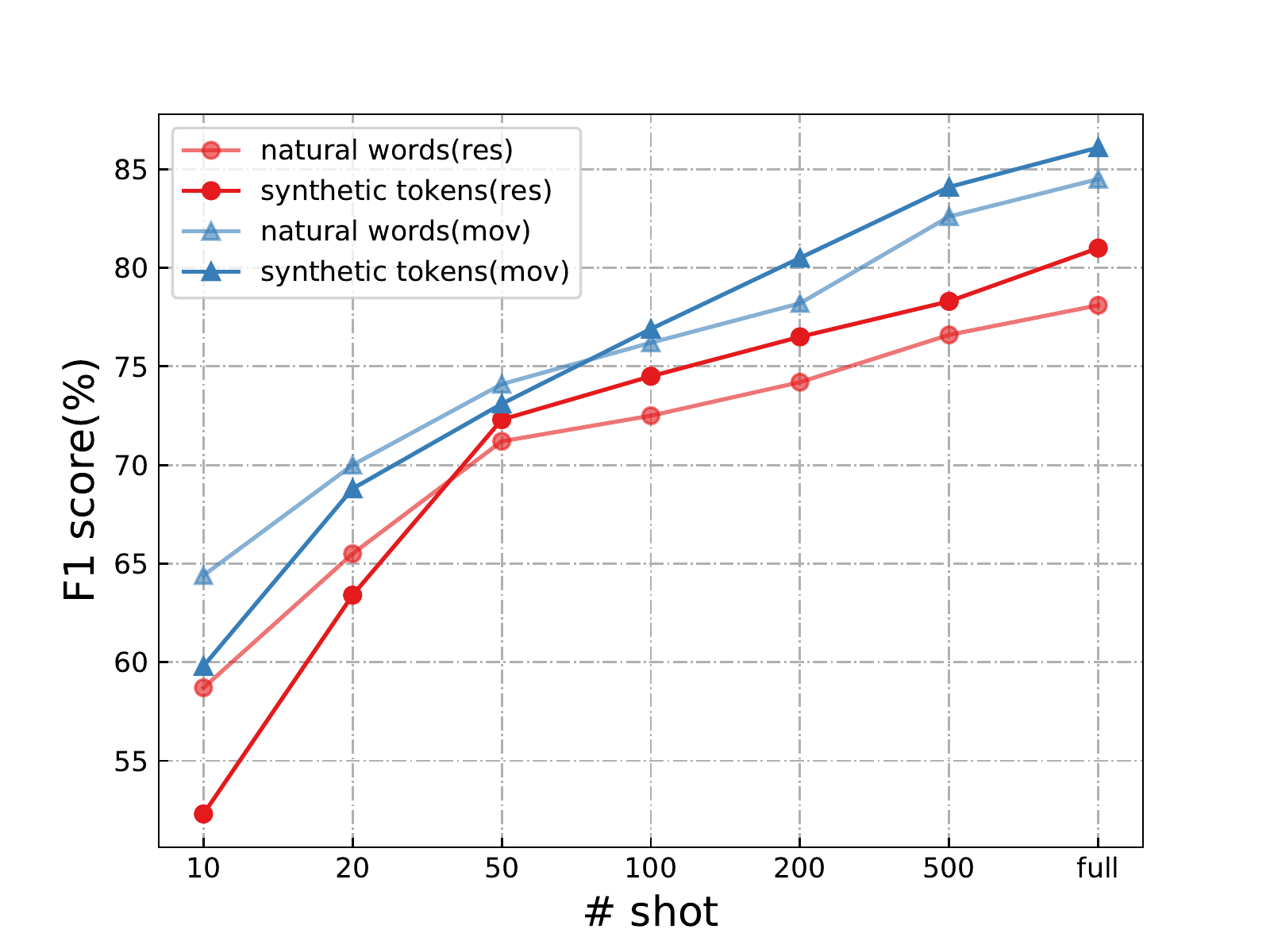}}
  \caption{F1 score(\%) on MIT Restaurant(res) and MIT Movie(mov) dataset with two type words strategies under different shot settings.}
  \label{fig:vocab}
  \vspace{-0.5cm}
\end{figure}

As illustrated in Figure \ref{fig:vocab}, we observe that: 1) The model performs worse in the few-shot scenarios when the synthetic tokens are used as entity types. 2) However, it is unexpected that the strategy of using synthetic tokens has superior performance than the other strategy in the rich-resource scenarios. We conjecture that 1) in the low-resource scenarios, the semantic information in the label words of the natural language form can help model stimulate latent knowledge hidden in PLMs and generate the entity words more accurately, while synthetic tokens are unable to do this. 2) However, when there are enough supervised signals, the synthetic tokens gradually learn the specific semantic relationship for a certain task, and the original semantics of label words in the natural language form would interfere with the specialization process of these words. 

\vspace{-0.2cm}

\section{Conclusion}
\vspace{-0.1cm}
In this paper, we reformulate the sequence labeling task as a generation problem and propose a multi-task instruction-based generative framework (InstructionNER) for low-resource NER. By constructing source sentences with descriptive task instructions and limited answer options, InstructionNER can make the best use of semantic knowledge learned by PLMs. Moreover, we further introduce two auxiliary tasks to enable the model to capture more boundary information of entities and semantic of types.
Experiments on several NER datasets show that our method consistently outperforms other baselines, proving its effectiveness in few-shot settings.


\bibliography{anthology,custom}

\begin{thebibliography}{31}
\expandafter\ifx\csname natexlab\endcsname\relax\def\natexlab#1{#1}\fi

\bibitem[{Brown et~al.(2020)Brown, Mann, Ryder, Subbiah, Kaplan, Dhariwal,
  Neelakantan, Shyam, Sastry, Askell, Agarwal, Herbert-Voss, Krueger, Henighan,
  Child, Ramesh, Ziegler, Wu, Winter, Hesse, Chen, Sigler, Litwin, Gray, Chess,
  Clark, Berner, McCandlish, Radford, Sutskever, and
  Amodei}]{NEURIPS2020_1457c0d6}
Tom Brown, Benjamin Mann, Nick Ryder, Melanie Subbiah, Jared~D Kaplan, Prafulla
  Dhariwal, Arvind Neelakantan, Pranav Shyam, Girish Sastry, Amanda Askell,
  Sandhini Agarwal, Ariel Herbert-Voss, Gretchen Krueger, Tom Henighan, Rewon
  Child, Aditya Ramesh, Daniel Ziegler, Jeffrey Wu, Clemens Winter, Chris
  Hesse, Mark Chen, Eric Sigler, Mateusz Litwin, Scott Gray, Benjamin Chess,
  Jack Clark, Christopher Berner, Sam McCandlish, Alec Radford, Ilya Sutskever,
  and Dario Amodei. 2020.
\newblock \href
  {https://proceedings.neurips.cc/paper/2020/file/1457c0d6bfcb4967418bfb8ac142f64a-Paper.pdf}
  {Language models are few-shot learners}.
\newblock In \emph{Advances in Neural Information Processing Systems},
  volume~33, pages 1877--1901. Curran Associates, Inc.

\bibitem[{Chen et~al.(2021)Chen, Zhang, Li, Xie, Deng, Tan, Huang, Si, and
  Chen}]{chen2021LightNER}
Xiang Chen, Ningyu Zhang, Lei Li, Xin Xie, Shumin Deng, Chuanqi Tan, Fei Huang,
  Luo Si, and Huajun Chen. 2021.
\newblock \href {http://arxiv.org/abs/2109.00720} {Lightner: A lightweight
  generative framework with prompt-guided attention for low-resource ner}.

\bibitem[{Chiu and Nichols(2016)}]{chiu-nichols-2016-named}
Jason~P.C. Chiu and Eric Nichols. 2016.
\newblock \href {https://doi.org/10.1162/tacl_a_00104} {Named entity
  recognition with bidirectional {LSTM}-{CNN}s}.
\newblock \emph{Transactions of the Association for Computational Linguistics},
  4:357--370.

\bibitem[{Cui et~al.(2021)Cui, Wu, Liu, Yang, and
  Zhang}]{cui-etal-2021-template}
Leyang Cui, Yu~Wu, Jian Liu, Sen Yang, and Yue Zhang. 2021.
\newblock \href {https://doi.org/10.18653/v1/2021.findings-acl.161}
  {Template-based named entity recognition using {BART}}.
\newblock In \emph{Findings of the Association for Computational Linguistics:
  ACL-IJCNLP 2021}, pages 1835--1845, Online. Association for Computational
  Linguistics.

\bibitem[{Dai et~al.(2020)Dai, Karimi, Hachey, and
  Paris}]{dai-etal-2020-effective}
Xiang Dai, Sarvnaz Karimi, Ben Hachey, and Cecile Paris. 2020.
\newblock \href {https://doi.org/10.18653/v1/2020.acl-main.520} {An effective
  transition-based model for discontinuous {NER}}.
\newblock In \emph{Proceedings of the 58th Annual Meeting of the Association
  for Computational Linguistics}, pages 5860--5870, Online. Association for
  Computational Linguistics.

\bibitem[{Devlin et~al.(2019)Devlin, Chang, Lee, and
  Toutanova}]{devlin-etal-2019-bert}
Jacob Devlin, Ming-Wei Chang, Kenton Lee, and Kristina Toutanova. 2019.
\newblock \href {https://doi.org/10.18653/v1/N19-1423} {{BERT}: Pre-training of
  deep bidirectional transformers for language understanding}.
\newblock In \emph{Proceedings of the 2019 Conference of the North {A}merican
  Chapter of the Association for Computational Linguistics: Human Language
  Technologies, Volume 1 (Long and Short Papers)}, pages 4171--4186,
  Minneapolis, Minnesota. Association for Computational Linguistics.

\bibitem[{Gao et~al.(2018)Gao, Galley, and
  Li}]{gao-etal-2018-neural-approaches}
Jianfeng Gao, Michel Galley, and Lihong Li. 2018.
\newblock \href {https://doi.org/10.18653/v1/P18-5002} {Neural approaches to
  conversational {AI}}.
\newblock In \emph{Proceedings of the 56th Annual Meeting of the Association
  for Computational Linguistics: Tutorial Abstracts}, pages 2--7, Melbourne,
  Australia. Association for Computational Linguistics.

\bibitem[{Gao et~al.(2021)Gao, Fisch, and Chen}]{gao2021making}
Tianyu Gao, Adam Fisch, and Danqi Chen. 2021.
\newblock \href {http://arxiv.org/abs/2012.15723} {Making pre-trained language
  models better few-shot learners}.

\bibitem[{Guo et~al.(2009)Guo, Xu, Cheng, and Li}]{guo2009named}
Jiafeng Guo, Gu~Xu, Xueqi Cheng, and Hang Li. 2009.
\newblock Named entity recognition in query.
\newblock In \emph{Proceedings of the 32nd international ACM SIGIR conference
  on Research and development in information retrieval}, pages 267--274.

\bibitem[{Hakkani-T{\"u}r et~al.(2016)Hakkani-T{\"u}r, T{\"u}r, Çelikyilmaz,
  Chen, Gao, Deng, and Wang}]{HakkaniTr2016MultiDomainJS}
Dilek~Z. Hakkani-T{\"u}r, G{\"o}khan T{\"u}r, Asli Çelikyilmaz,
  Yun-Nung~(Vivian) Chen, Jianfeng Gao, Li~Deng, and Ye-Yi Wang. 2016.
\newblock Multi-domain joint semantic frame parsing using bi-directional
  rnn-lstm.
\newblock In \emph{INTERSPEECH}.

\bibitem[{Han et~al.(2021)Han, Zhang, Ding, Gu, Liu, Huo, Qiu, Yao, Zhang,
  Zhang, Han, Huang, Jin, Lan, Liu, Liu, Lu, Qiu, Song, Tang, Wen, Yuan, Zhao,
  and Zhu}]{han2021pretrained}
Xu~Han, Zhengyan Zhang, Ning Ding, Yuxian Gu, Xiao Liu, Yuqi Huo, Jiezhong Qiu,
  Yuan Yao, Ao~Zhang, Liang Zhang, Wentao Han, Minlie Huang, Qin Jin, Yanyan
  Lan, Yang Liu, Zhiyuan Liu, Zhiwu Lu, Xipeng Qiu, Ruihua Song, Jie Tang,
  Ji-Rong Wen, Jinhui Yuan, Wayne~Xin Zhao, and Jun Zhu. 2021.
\newblock \href {http://arxiv.org/abs/2106.07139} {Pre-trained models: Past,
  present and future}.

\bibitem[{Huang et~al.(2020)Huang, Li, Subudhi, Jose, Balakrishnan, Chen, Peng,
  Gao, and Han}]{huang2020fewshot}
Jiaxin Huang, Chunyuan Li, Krishan Subudhi, Damien Jose, Shobana Balakrishnan,
  Weizhu Chen, Baolin Peng, Jianfeng Gao, and Jiawei Han. 2020.
\newblock \href {http://arxiv.org/abs/2012.14978} {Few-shot named entity
  recognition: A comprehensive study}.

\bibitem[{Lewis et~al.(2020)Lewis, Liu, Goyal, Ghazvininejad, Mohamed, Levy,
  Stoyanov, and Zettlemoyer}]{lewis-etal-2020-bart}
Mike Lewis, Yinhan Liu, Naman Goyal, Marjan Ghazvininejad, Abdelrahman Mohamed,
  Omer Levy, Veselin Stoyanov, and Luke Zettlemoyer. 2020.
\newblock \href {https://doi.org/10.18653/v1/2020.acl-main.703} {{BART}:
  Denoising sequence-to-sequence pre-training for natural language generation,
  translation, and comprehension}.
\newblock In \emph{Proceedings of the 58th Annual Meeting of the Association
  for Computational Linguistics}, pages 7871--7880, Online. Association for
  Computational Linguistics.

\bibitem[{Liu et~al.(2021)Liu, Fu, Tan, Chen, Zhang, Huang, and
  Gao}]{liu-etal-2021-noisy-labeled}
Kun Liu, Yao Fu, Chuanqi Tan, Mosha Chen, Ningyu Zhang, Songfang Huang, and
  Sheng Gao. 2021.
\newblock \href {https://doi.org/10.18653/v1/2021.naacl-main.269}
  {Noisy-labeled {NER} with confidence estimation}.
\newblock In \emph{Proceedings of the 2021 Conference of the North American
  Chapter of the Association for Computational Linguistics: Human Language
  Technologies}, pages 3437--3445, Online. Association for Computational
  Linguistics.

\bibitem[{Liu et~al.(2019)Liu, Meng, Zhang, Xu, Chen, and
  Zhou}]{liu-etal-2019-gcdt}
Yijin Liu, Fandong Meng, Jinchao Zhang, Jinan Xu, Yufeng Chen, and Jie Zhou.
  2019.
\newblock \href {https://doi.org/10.18653/v1/P19-1233} {{GCDT}: A global
  context enhanced deep transition architecture for sequence labeling}.
\newblock In \emph{Proceedings of the 57th Annual Meeting of the Association
  for Computational Linguistics}, pages 2431--2441, Florence, Italy.
  Association for Computational Linguistics.

\bibitem[{Ma and Hovy(2016)}]{ma-hovy-2016-end}
Xuezhe Ma and Eduard Hovy. 2016.
\newblock \href {https://doi.org/10.18653/v1/P16-1101} {End-to-end sequence
  labeling via bi-directional {LSTM}-{CNN}s-{CRF}}.
\newblock In \emph{Proceedings of the 54th Annual Meeting of the Association
  for Computational Linguistics (Volume 1: Long Papers)}, pages 1064--1074,
  Berlin, Germany. Association for Computational Linguistics.

\bibitem[{Metke-Jimenez and Karimi(2016)}]{metke2016concept}
Alejandro Metke-Jimenez and Sarvnaz Karimi. 2016.
\newblock Concept identification and normalisation for adverse drug event
  discovery in medical forums.
\newblock In \emph{BMDID@ ISWC}. Citeseer.

\bibitem[{Mishra et~al.(2021)Mishra, Khashabi, Baral, and
  Hajishirzi}]{mishra2021crosstask}
Swaroop Mishra, Daniel Khashabi, Chitta Baral, and Hannaneh Hajishirzi. 2021.
\newblock \href {http://arxiv.org/abs/2104.08773} {Cross-task generalization
  via natural language crowdsourcing instructions}.

\bibitem[{Moll{\'a} et~al.(2006)Moll{\'a}, van Zaanen, and
  Smith}]{molla-etal-2006-named}
Diego Moll{\'a}, Menno van Zaanen, and Daniel Smith. 2006.
\newblock \href {https://aclanthology.org/U06-1009} {Named entity recognition
  for question answering}.
\newblock In \emph{Proceedings of the Australasian Language Technology Workshop
  2006}, pages 51--58, Sydney, Australia.

\bibitem[{Raffel et~al.(2020)Raffel, Shazeer, Roberts, Lee, Narang, Matena,
  Zhou, Li, and Liu}]{raffel2020exploring}
Colin Raffel, Noam Shazeer, Adam Roberts, Katherine Lee, Sharan Narang, Michael
  Matena, Yanqi Zhou, Wei Li, and Peter~J. Liu. 2020.
\newblock \href {http://arxiv.org/abs/1910.10683} {Exploring the limits of
  transfer learning with a unified text-to-text transformer}.

\bibitem[{Ratinov and Roth(2009)}]{ratinov-roth-2009-design}
Lev Ratinov and Dan Roth. 2009.
\newblock \href {https://aclanthology.org/W09-1119} {Design challenges and
  misconceptions in named entity recognition}.
\newblock In \emph{Proceedings of the Thirteenth Conference on Computational
  Natural Language Learning ({C}o{NLL}-2009)}, pages 147--155, Boulder,
  Colorado. Association for Computational Linguistics.

\bibitem[{Ritter et~al.(2012)Ritter, Etzioni, and Clark}]{ritter2012open}
Alan Ritter, Oren Etzioni, and Sam Clark. 2012.
\newblock Open domain event extraction from twitter.
\newblock In \emph{Proceedings of the 18th ACM SIGKDD international conference
  on Knowledge discovery and data mining}, pages 1104--1112.

\bibitem[{Schick and Sch{\"u}tze(2021)}]{schick-schutze-2021-exploiting}
Timo Schick and Hinrich Sch{\"u}tze. 2021.
\newblock \href {https://doi.org/10.18653/v1/2021.eacl-main.20} {Exploiting
  cloze-questions for few-shot text classification and natural language
  inference}.
\newblock In \emph{Proceedings of the 16th Conference of the European Chapter
  of the Association for Computational Linguistics: Main Volume}, pages
  255--269, Online. Association for Computational Linguistics.

\bibitem[{Strakov{\'a} et~al.(2019)Strakov{\'a}, Straka, and
  Hajic}]{strakova-etal-2019-neural}
Jana Strakov{\'a}, Milan Straka, and Jan Hajic. 2019.
\newblock \href {https://doi.org/10.18653/v1/P19-1527} {Neural architectures
  for nested {NER} through linearization}.
\newblock In \emph{Proceedings of the 57th Annual Meeting of the Association
  for Computational Linguistics}, pages 5326--5331, Florence, Italy.
  Association for Computational Linguistics.

\bibitem[{Strubell et~al.(2017)Strubell, Verga, Belanger, and
  McCallum}]{strubell-etal-2017-fast}
Emma Strubell, Patrick Verga, David Belanger, and Andrew McCallum. 2017.
\newblock \href {https://doi.org/10.18653/v1/D17-1283} {Fast and accurate
  entity recognition with iterated dilated convolutions}.
\newblock In \emph{Proceedings of the 2017 Conference on Empirical Methods in
  Natural Language Processing}, pages 2670--2680, Copenhagen, Denmark.
  Association for Computational Linguistics.

\bibitem[{Sun et~al.(2021)Sun, Zheng, Hao, and Qiu}]{sun2021nspbert}
Yi~Sun, Yu~Zheng, Chao Hao, and Hangping Qiu. 2021.
\newblock \href {http://arxiv.org/abs/2109.03564} {Nsp-bert: A prompt-based
  zero-shot learner through an original pre-training task--next sentence
  prediction}.

\bibitem[{Wei et~al.(2021)Wei, Bosma, Zhao, Guu, Yu, Lester, Du, Dai, and
  Le}]{wei2021finetuned}
Jason Wei, Maarten Bosma, Vincent~Y. Zhao, Kelvin Guu, Adams~Wei Yu, Brian
  Lester, Nan Du, Andrew~M. Dai, and Quoc~V. Le. 2021.
\newblock \href {http://arxiv.org/abs/2109.01652} {Finetuned language models
  are zero-shot learners}.

\bibitem[{Wiseman and Stratos(2019)}]{wiseman-stratos-2019-label}
Sam Wiseman and Karl Stratos. 2019.
\newblock \href {https://doi.org/10.18653/v1/P19-1533} {Label-agnostic sequence
  labeling by copying nearest neighbors}.
\newblock In \emph{Proceedings of the 57th Annual Meeting of the Association
  for Computational Linguistics}, pages 5363--5369, Florence, Italy.
  Association for Computational Linguistics.

\bibitem[{Yan et~al.(2021)Yan, Gui, Dai, Guo, Zhang, and
  Qiu}]{yan-etal-2021-unified-generative}
Hang Yan, Tao Gui, Junqi Dai, Qipeng Guo, Zheng Zhang, and Xipeng Qiu. 2021.
\newblock \href {https://doi.org/10.18653/v1/2021.acl-long.451} {A unified
  generative framework for various {NER} subtasks}.
\newblock In \emph{Proceedings of the 59th Annual Meeting of the Association
  for Computational Linguistics and the 11th International Joint Conference on
  Natural Language Processing (Volume 1: Long Papers)}, pages 5808--5822,
  Online. Association for Computational Linguistics.

\bibitem[{Yang and Katiyar(2020)}]{yang2020simple}
Yi~Yang and Arzoo Katiyar. 2020.
\newblock \href {http://arxiv.org/abs/2010.02405} {Simple and effective
  few-shot named entity recognition with structured nearest neighbor learning}.

\bibitem[{Ziyadi et~al.(2020)Ziyadi, Sun, Goswami, Huang, and
  Chen}]{ziyadi2020examplebased}
Morteza Ziyadi, Yuting Sun, Abhishek Goswami, Jade Huang, and Weizhu Chen.
  2020.
\newblock \href {http://arxiv.org/abs/2008.10570} {Example-based named entity
  recognition}.

\end{thebibliography}
\bibliographystyle{acl_natbib}

\renewcommand\thesection{\Alph{section}}

\appendix

\section{Implementation  Details}
\label{sec:appendix}
In this paper, we use T5 as the backbone PLM in our experiments. There are no extra parameters are introduced, so the total parameters of our model is as same as original T5. Specifically, the number of parameters of T5-base\footnote{\url{https://huggingface.co/t5-base}}, T5-large\footnote{\url{https://huggingface.co/t5-large}} and T5-v1.1\footnote{\url{https://huggingface.co/google/t5-v1_1-large}} are 220M/770M/770M respectively. For all experiments, we train and test our model on a single V100 GPU. On average, each model required 2 GPU-hours to train. The deep learning frameworks used in this paper are Pytorch==1.7.1 and transformers==4.3.3.

\section{Dataset Statistics}

The statistical information as shown in Table 4.

\begin{table}[htbp]
    \centering
    \resizebox{75mm}{12mm}{
    \begin{tabular}{c|c|c}
        \hline
        \textbf{Dataset} & \textbf{Sample Account (train / test)} & \textbf{Entity Type}\\
        
        \hline
        CoNLL03 & 14986 / 3684 & 4\\
        
        \hline
        ATIS & 4478 / 893 & 79\\
        
        \hline
        MIT\_movie & 9775 / 2443 &12\\
        
        \hline
        MIT\_restaurant & 7660 / 1521& 8 \\
        
        \hline
        Ontonotes5.0 & 115812 / 12217& 18\\

        \hline
    \end{tabular}}
    \vspace{-0.3cm}
    \caption{Statistics of mentioned NER datasets}
    \label{tab:datas example}
\vspace{-0.7cm}
\end{table}

\section{Performance on Relatively Abundant Settings}
We conduct the experiments on MIT Movie and MIT Restaurant under the setting of 100/200/500-shot. As shown in Table \ref{tab:appendix}, when the data size grows to 100/200/500-shot for each entity type, the performance of InstructionNER still grows but with a relatively small increment, and even underperforms the LightNER baseline in the end.
It reveals that our InstructionNER model focuses more on learning the general knowledge and is trained in the form of meta-learning. Thus, it shows high efficiency in the data-scarce scenarios. However, when the data is sufficient, the generality may hamper the model's specialization so that adding extra samples shows limited performance improvement. We also discuss the similar conclusion in Section \ref{sec:label_semantic_analysis}.

\begin{table}[htbp]
  \centering
  \resizebox{75mm}{28mm}{
  \begin{tabular}{lcccccc}
    \toprule
    \multirow{2}{*}{Methods}&\multicolumn{3}{c}{MIT Movie}&\multicolumn{3}{c}{MIT Restaurant} \\
    \cmidrule(lr){2-4}
    \cmidrule(lr){5-7}
    &100&200&500&100&200&500 \\
    \midrule
    \multicolumn{1}{c}{\textit{Baselines}} \\
    LC-BERT&50.7&59.3&74.4&53.5&57.4&61.3 \\
    LC-BART&47.5&54.2&64.1&52.2&56.3&60.2 \\
    Template&56.3&62.0&74.9&60.1&62.8&65.0 \\
    BARTNER$^*$&70.1&74.6&82.6&65.3&74.4&75.7\\
    LightNER&\underline{\textbf{78.0}}&\underline{\textbf{80.6}}&\underline{\textbf{84.8}}&\underline{70.8}&\underline{75.5}&\underline{\textbf{80.2}}\\
    \midrule
  
    \multicolumn{1}{c}{\textit{Our implementations}} \\

    InstructionNER&
   
    76.2&
    78.2&
    \underline{82.6}&
    
    72.5&
    74.2&
    \underline{76.6}
    \\

    InstructionNER{\scriptsize +ET}&
   
    \underline{76.9}&
    77.7&
    82.2&
    
    73.3&
    \underline{\textbf{76.0}}&
    76.1
    \\

    InstructionNER{\scriptsize +EE}&
    
    76.2&
    \underline{78.4}&
    82.5&
    
    \underline{\textbf{73.4}}&
    75.9&
    76.0
    \\

    InstructionNER{\scriptsize +ET,EE}&
    
    74.3&
    \underline{78.4}&
    82.3&
    
    72.7&
    75.5&
    \underline{76.6}
    \\
    \bottomrule

  \end{tabular}}
  \caption{F1 score(\%) on MIT datasets under 100/200/500 shot settings. \textbf{Bold} numbers indicate the best performance across baselines and our implementations, and \underline{underline} indicates the best performance in baselines or our implementations. `` * '' indicates our reproduction results.}
  \label{tab:appendix}
  \vspace{-0.5cm}
  
\end{table}

\begin{figure}
  \centering
  \resizebox{\columnwidth}{!}{
  \includegraphics{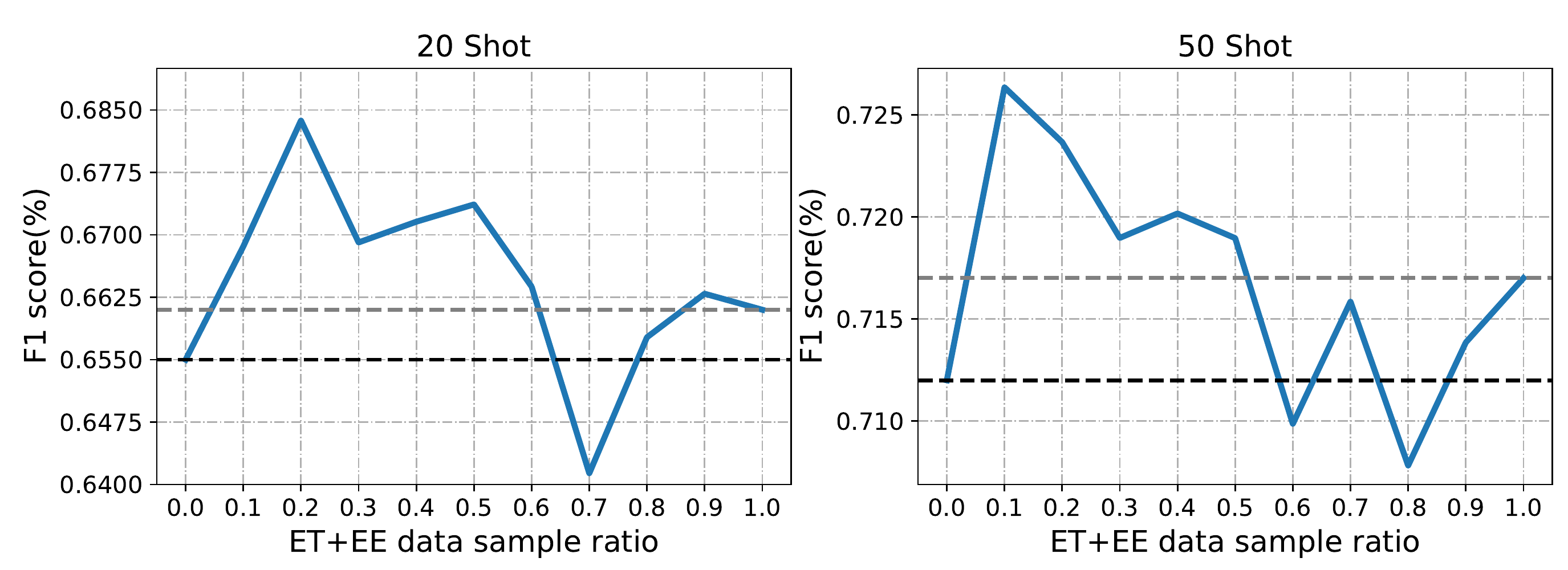}}
  \caption{F1 score(\%) on MIT Restaurant dataset with different ET+EE data sample ratios under 20/50 shot settings.}
  \label{fig:task-ratio}
  \vspace{-0.5cm}
\end{figure}

\section{The Ratio of Auxiliary Tasks}
We notice that the performance is dropped when we combine the ET and EE tasks. We conjecture the reason is that the data size ratio between the main task and the auxiliary tasks is reduced to 1:2 when introducing ET and EE simultaneously, which means our model will focus more on auxiliary tasks but less on the main task, leading to a performance drop on the main NER task. 
To validate this, we sample different ratios of auxiliary data and conduct experiments to evaluate the F1 score. As illustrated in Figure \ref{fig:task-ratio}, when the data of auxiliary tasks is less or equal than the main task (i.e., the sample ratio is between 0.1 and 0.5), InstructionNER can achieve a better F1 score than the one trained without auxiliary tasks (i.e., the sample ratio is 0) or with full auxiliary tasks data (i.e., the sample ratio is 1). When auxiliary tasks data grows larger than the main task (i.e., the sample ratio > 0.5), the performance drops in different degrees and sometimes even underperforms the one without auxiliary tasks.

\end{document}